\documentclass[11pt,a4paper]{article}
\usepackage[hyperref]{emnlp2018}
\usepackage{times}
\usepackage{latexsym}
\usepackage{url}
\usepackage{amssymb}
\usepackage{url}
\usepackage{amsmath}
\usepackage{latexsym}
\usepackage{balance}
\usepackage{booktabs}
\usepackage{array,multirow,graphicx}
\usepackage{tikz}
\usepackage[ruled]{algorithm2e}
\usetikzlibrary{shapes}
\usetikzlibrary{fit,positioning,calc,backgrounds}
\usetikzlibrary{decorations.pathreplacing}
\usepackage{array,multirow,graphicx}

\aclfinalcopy 

\title{Concurrent Learning of Semantic Relations}

\author{Georgios Balikas \\
  Kelkoo Group, Grenoble, France \\
  {\tt geompalik@hotmail.com} \\\And
  Ga\"el Dias \\
  University of Caen Normandy, France \\
  {\tt gael.dias@unicaen.fr} \\ \AND 
  Rumen Moraliyski\\
  Kodar Ltd, Plovdiv, Bulgaria  \\
  {\tt rumen.moraliyski@gmail.com} \\ \And
  Massih-Reza Amini \\
  University of Grenoble Alps,  France\\
  {\tt massih.amini@imag.fr} \\
  }

\date{}

\begin{document}
\maketitle

\begin{abstract}
Discovering whether words are semantically related and identifying the specific semantic relation that holds between them is of crucial importance for NLP as it is essential for tasks like query expansion in IR. Within this context, different methodologies have been proposed that either exclusively focus on a single lexical relation (e.g. hypernymy vs. random) or learn specific classifiers capable of identifying multiple semantic relations (e.g. hypernymy vs. synonymy vs. random). In this paper, we propose another way to look at the problem that relies on the multi-task learning paradigm. In particular, we want to study whether the learning process of a given semantic relation (e.g. hypernymy) can be improved by the concurrent learning of another semantic relation (e.g. co-hyponymy). Within this context, we particularly examine the benefits of semi-supervised learning where the training of a prediction function is performed over few labeled data jointly with many unlabeled ones. Preliminary results based on simple learning strategies and state-of-the-art distributional feature representations show that concurrent learning can lead to improvements in a vast majority of tested situations.      
\end{abstract}


\section{Introduction}

The ability to automatically identify semantic relations is an important issue for Information Retrieval (IR) and Natural Language Processing (NLP) applications such as question answering \cite{Dong2017}, query expansion \cite{kathuria2017comprehensive}, or text summarization \cite{gambhir2017recent}. Semantic relations embody a large number of symmetric and asymmetric linguistic phenomena such as synonymy (bike $\leftrightarrow$ bicycle), co-hyponymy (bike $\leftrightarrow$ scooter), hypernymy (bike $\rightarrow$ tandem) or meronymy (bike $\rightarrow$ chain), but more can be enumerated \cite{Vylomova2016}. 

Most approaches focus on modeling a single semantic relation and consist in deciding whether a given relation $r$ holds between a pair of words ($x$,$y$) or not. Within this context, the vast majority of efforts \cite{Snow2004,Roller2014,Shwartz2016,Nguyen2017} concentrate on hypernymy which is the key organization principle of semantic memory, but studies exist on antonymy \cite{Nguyen2017a}, meronymy \cite{Glavas2017} and co-hyponymy \cite{Weeds2014}. Another research direction consists in dealing with multiple semantic relations at a time and can be defined as deciding which semantic relation $r_i$ (if any) holds between a pair of words $(x,y)$. This multi-class problem is challenging as it is known that distinguishing between different semantic relations (e.g. synonymy and hypernymy) is difficult \cite{Shwartz2016}. Within the CogALex-V shared task\footnote{\url{https://sites.google.com/site/cogalex2016/home/shared-task}} which aims at tackling synonymy, antonymy, hypernymy and meronymy as a multi-class problem, best performing systems are proposed by \cite{Shwartz2016a} and \cite{Attia2016}.  

In this paper, we propose another way to look at the problem of semantic relation identification based on the following findings. First, symmetric similarity measures, which capture synonymy \cite{Kiela2015} have shown to play an important role in hypernymy detection \cite{Santus2017}. Second, \cite{Yu2015} show that learning term embeddings that take into account co-hyponymy similarity improves supervised hypernymy identification. As a consequence, we propose to study whether the learning process of a given semantic relation can be improved by the concurrent learning of another relation, where semantic relations are either synonymy, co-hyponymy or hypernymy. For that purpose, we propose a multi-task learning strategy using a hard parameter sharing neural network model that takes as input a learning word pair $(x,y)$ encoded as a feature vector representing the concatenation\footnote{Best configuration reported in \cite{Shwartz2016} for standard non path-based supervised learning.} of the respective word embeddings of $x$ and $y$ noted $\small \overrightarrow{x}\oplus\overrightarrow{y}$. The intuition behind our experiments is that if the tasks are correlated, the neural network should improve its generalization ability by taking into account the shared information.

In parallel, we propose to study the generalization ability of the multi-task learning model based on a limited set of labeled word pairs and a large number of unlabeled samples, i.e. following a semi-supervised paradigm. As far as we know, most related works rely on the existence of a huge number of validated word pairs present in knowledge databases (e.g. WordNet) to perform the supervised learning process. However, such resources may not be available for specific languages or domains. Moreover, it is unlikely that human cognition and its generalization capacity rely on the equivalent number of positive examples. As such, semi-supervised learning proposes a much more interesting framework where unlabeled word pairs can massively\footnote{Even though with some error rate.} be obtained through selected lexico-syntactic patterns \cite{Hearst1992} or paraphrase alignments \cite{dias2010}. To test our hypotheses, we propose a self-learning strategy where confidently tagged unlabeled word pairs are iteratively added to the labeled dataset.  

Preliminary results based on simple (semi-supervised) multi-task learning models with state-of-the-art word pairs representations (i.e. concatenation of GloVe \cite{pennington2014glove} word embeddings) over the gold standard dataset ROOT9 \cite{Santus2016} and the RUMEN dataset proposed in this paper show that classification improvements can be obtained for a wide range of tested configurations.


\section{Related Work}

Whether semantic relation identification has been tackled as a one-class or a multi-class problem, two main approaches have been addressed to capture the semantic links between two words $(x,y)$: pattern-based and distributional. Pattern-based (a.k.a. path-based) methods base their decisions on the analysis of the lexico-syntactic patterns (e.g. X {\it such as} Y) that connect the joint occurrences of $x$ and $y$. Within this context, earlier works have been proposed by \cite{Hearst1992} (unsupervised) and \cite{Snow2004} (supervised) to detect hypernymy. However, this approach suffers from sparse coverage and benefits precision over recall. To overcome these limitations, recent one-class studies on hypernymy \cite{Shwartz2016} and antonymy \cite{Nguyen2017a}, as well as multi-class approaches \cite{Shwartz2016a} have been focusing on representing dependency patterns as continuous vectors using long short-term memory (LSTM) networks. Within this context, successful results have been evidenced but \cite{Shwartz2016,Nguyen2017a} also show that the combination of pattern-based methods with the distributional approach greatly improves performance. 

In distributional methods, the decision whether $x$ is within a semantic relation with $y$ is based on the distributional representation of these words following the distributional hypothesis \cite{Harris1954}, i.e. on the separate contexts of $x$ and $y$. Earlier works developed symmetric \cite{dias2010} and asymmetric \cite{Kotlerman2010} similarity measures based on discrete representation vectors, followed by numerous supervised learning strategies for a wide range of semantic relations \cite{Baroni2012,Roller2014,Weeds2014}, where word pairs are encoded as the concatenation of the constituent words representations ($\small \overrightarrow{x}\oplus\overrightarrow{y}$) or their vector difference ($\small \overrightarrow{x} - \overrightarrow{y}$). More recently, attention has been focusing on identifying semantic relations using neural language embeddings, as such semantic spaces encode linguistic regularities \cite{Mikolov2013}. Within this context, \cite{Vylomova2016} proposed an exhaustive study for a wide range of semantic relations and showed that under suitable supervised training, high performance can be obtained. However, \cite{Vylomova2016} also showed that some relations such as hypernymy are more difficult to model than others. As a consequence, new proposals have appeared that tune word embeddings for this specific task, where hypernyms and hyponyms should be closed to each other in the semantic space \cite{Yu2015,Nguyen2017}.

\begin{figure}[h!t]
\begin{center}
\includegraphics[scale=0.15]{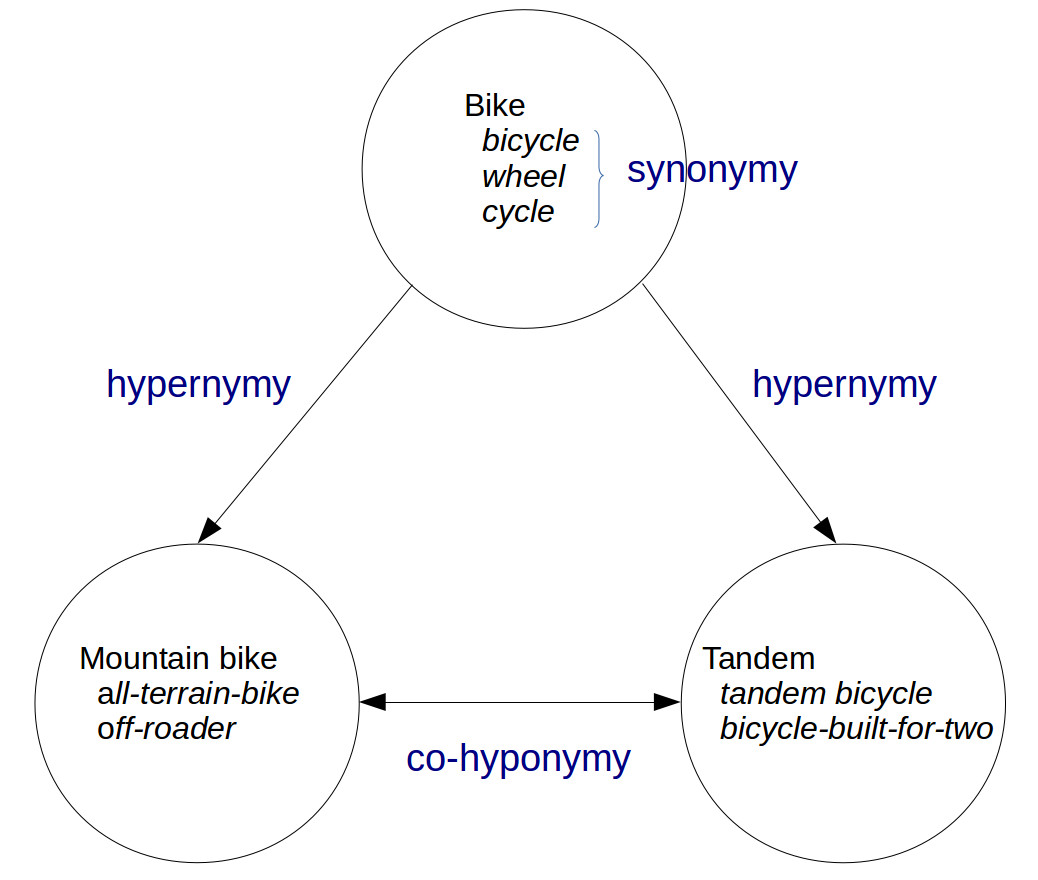}
\end{center}
\caption{\label{onto} Basic unit of a taxonomic structure.}
\end{figure}

In this paper, we propose an attempt to deal with semantic relation identification based on a multi-task strategy, as opposed to previous one-class and multi-class approaches. Our main scope is to analyze whether a link exists between the learning process of related semantic relations. The closest approach to ours is proposed by \cite{Attia2016}, which develops a multi-task convolutional neural network for multi-class semantic relation classification supported by relatedness classification. As such, it can be seen as a domain adaptation problem. Within the scope of our paper, we aim at studying semantic inter-relationships at a much finer grain and understanding the cognitive links that may exist between synonymy, co-hyponymy and hypernymy, that form the backbone of any taxonomic structure as illustrated in Figure \ref{onto}. For this first attempt, we follow the distributional approach as in \cite{Attia2016}, although we are aware that improvements may be obtained by the inclusion of pattern-based representations\footnote{This issue is out of the scope of this paper.}. Moreover, we propose the first attempt\footnote{As far as we know.} to deal with semantic relation identification based on a semi-supervised approach, thus avoiding the pre-existence of a large number of training examples. As a consequence, we aim at providing a more natural learning framework where only a few labeled examples are initially provided and massively gathered related word pairs iteratively improve learning.

\section{Methodology}


\subsection{Multi-task with hard parameter sharing}
As discussed in \cite{bingel2017identifying}, not every task combination is beneficial. But, concurrent learning of tasks that have cognitive similarities is often beneficial. We may hypothesize that recognizing the different semantic relations that hold between words can benefit classification models across similar tasks. For instance, learning that {\em bike} is the hypernym of {\em mountain bike} should help while classifying {\em mountain bike} and {\em tandem bicycle} as co-hyponyms, as it is likely that {\em tandem bicycle} shares some relation with {\em bike}. To test this hypothesis, we propose to use a multi-task learning approach. Multi-task learning \cite{caruana1998multitask} has empirically been validated and has shown to be effective in a variety of NLP tasks ranging from sentiment analysis, part-of-speech tagging and text parsing \cite{plank2016coling,bingel2017identifying}. The hope is that by jointly learning the decision functions for related tasks, one can achieve better performance. It may be first due to knowledge transfer across tasks that is achieved either in the form of learning more robust representations or due to the use of more data. Second, it has been argued that multi-task learning can act as a regularization process thus preventing from overfitting by requiring competitive performance across different tasks \cite{caruana1998multitask}. 

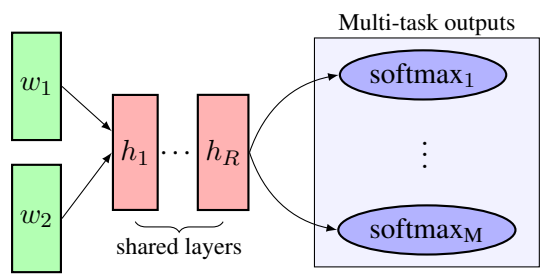
\begin{figure}[h!t]\centering
\begin{tikzpicture}
  \tikzstyle{connect}=[draw, -latex]
  \node[rectangle, fill=green!30, draw, thick, inner xsep=0.1cm, inner ysep=0.60cm, ] (hidden) at (0,0) {$w_1$};
  \node[rectangle, fill=green!30, draw, thick, inner xsep=0.1cm, inner ysep=0.60cm,  below  = 0.3cm  of hidden ] (hidden3)  {$w_2$};

  \node[rectangle, fill=red!30, draw, thick, inner xsep=0.1cm, inner ysep=0.60cm, ] (hidden2) at  ($(hidden3)!0.5!(hidden)+(1.3,0)$) {$h_1$};
  \node[rectangle, fill=red!30, draw, thick, inner xsep=0.1cm, inner ysep=0.60cm, right = 0.5cm of hidden2] (hidden5)   {$h_R$};

  \node[ellipse, fill=blue!30, draw, thick, inner sep=2pt, above right = 0.01cm and 1.5cm of hidden5] (o1) { softmax$_1$};
  \node[ellipse, fill=blue!30, draw, thick, inner sep=2pt, below right = 0.01cm and 1.5cm of hidden5] (o2) { softmax$_\text{M}$};
  \begin{scope}[on background layer]
	  \node [draw,  fill=blue!5, fit= (o1) (o2), inner xsep=0.3cm, inner ysep=0.15cm, label={[xshift=0mm, yshift=-1mm]above : \small Multi-task outputs}] {};
  \end{scope}
  \path 
  (hidden.east) edge [connect]  (hidden2)
  (hidden3.east) edge [connect] (hidden2.west)
  (hidden5.east) edge [connect, bend left]  (o1.west)
  (hidden5.east) edge [connect, bend right]  (o2.west)
  (o1) -- node[auto=false,rotate=90]{\ldots} (o2)
  (hidden2) -- node[auto=false]{\ldots} (hidden5);
;
\draw [decorate,decoration={brace,amplitude=4pt,mirror,raise=4pt,},yshift=0pt]
(hidden2.south) -- (hidden5.south) node [black,midway, yshift=-5mm] {\footnotesize shared layers};
\end{tikzpicture}
\caption{The multi-task learning architecture is built on top of a feed-forward neural network, where the layers $h_1 \cdots h_R$ are shared across tasks while the output layers softmax$_1 \cdots$ softmax$_M$ are task-dependent.} \label{fig:nn-archi}
\end{figure}

In this paper, we propose to use a multi-task learning algorithm that relies on hard parameter sharing\footnote{We use a simple architecture as our primary objective is to validate our initial hypotheses and not necessarily focus on overall performance.}. The idea is that the shared parameters (e.g. word representations or weights of some hidden layers) can benefit the performance of all tasks learned concurrently if the tasks are related. In particular, we propose a hard parameter sharing architecture based on a feed-forward neural network (NN) to perform the classification step. The NN architecture is illustrated in Figure \ref{fig:nn-archi}.

The input of the network is the concatenation of the word embeddings of the word pairs followed by a series of non-linear hidden layers. Then, a number of softmax layers gives the network predictions. Here, a softmax layer corresponds to a task, and concurrently learning $M$ tasks requires $M$ separate output softmax layers. The efficiency of hard parameter sharing architectures relies on the fact that the first layers that are shared are tuned by back-propagating the classification errors of every task. That way, the architecture uses the datasets of all tasks, instead of just one at a time. In Algorithm \ref{algo:multitask-training}, we detail the training protocol. Note that the different tasks learned by the NN share the same weights as batches are randomly sampled from their corresponding datasets\footnote{Automatically learning different weights for the tasks and self-adjusting them for the sake of overall performance is a possible future research direction.}. 

\begin{algorithm}
\small
 \KwData{Labeled words pairs $\mathcal{L}^i$ for each of the $M$ tasks, batch size b, epochs }
epoch = 1 \;
\While{epoch $<$ epochs}{
\For{$i = 0;\ i < M;\ i = i + 1$}{
    Randomly select a batch of size $b$ for task $i$ \;
    Update the parameters of the neural network architecture 
    according to the errors observed for the batch\;
    Calculate the performance on the validation set of task $i$. 
    }
}
 \caption{Multi-task Training Process}\label{algo:multitask-training}
\end{algorithm}

\subsection{Semi-supervision via self-learning}
Semi-supervised learning approaches have shown to perform well in a variety of tasks such as text classification and text summarization \cite{AminiG02, chapelle2009semi}. As in the supervised learning framework, we assume that we are given access to a set $\mathcal{L}=\{(w_i,w_i',rel)\}_{i=1}^{i=K}$ that consists of $K$ pairs of words labeled according to the relationship $rel$. Complementary to that, we also assume to have access to a set of $K'$ words pairs $\mathcal{U}=\{(w_i,w_i')\}_{i=1}^{i=K'}$ distinct from those of $\mathcal{L}$, and totally unlabeled. The challenge in this setting is to surpass the performance of classification models trained exclusively on $\mathcal{L}$ by using the available data in $\mathcal{U}$. While several methods have been proposed, we opt to use self-learning as the semi-supervised algorithm, as it is one of the simplest approaches\footnote{The main objective of this paper is to verify the benefits of semi-supervised multi-task learning for our task and not to tune complex solutions that can reach high performance. This remains a future work.}. The central idea behind self-learning is to train a learner on the set $\mathcal{L}$, and then progressively expand $\mathcal{L}$, by pseudo-labeling $N$ pairs within $\mathcal{U}$, for which the current prediction function is the most confident and adding them to $\mathcal{L}$. This process is repeated until no more pairs are available in $\mathcal{U}$ or, that the performance on a validation set degrades due to the newly-added possibly noisy examples. Algorithm \ref{algo:self-learning} details this process. 

\begin{algorithm}
\small
 \KwData{Word pairs: labeled $\mathcal{L}$, unlabeled $\mathcal{U}$, validation $\mathcal{V}$; integer $N$}
$\mathcal{L}_0$ = $\mathcal{L}$, $\mathcal{U}_0$ = $\mathcal{U}$ \;
Train classifier $C$ using $\mathcal{L}_0$ \;
$V_0: $ Performance of  $C$ on $\mathcal{V}$ \;
Set $t=0$\;
 \While{$\text{Size}(\mathcal{U}_t)>0$ and $V_{t}$ $\geqslant$ $V_0$  }{
 Get probability scores  $p$ of $C$ on $\mathcal{U}_t$ \;
 $\text{pseudo\_labeled} (N)= \arg\max(p)$, \textbf{stratified  wrt $\mathcal{L}_0$ } \;
 t = t + 1\;
 $\mathcal{L}_t = \mathcal{L}_{t-1} + \text{pseudo\_labeled}$ \;
 $\mathcal{U}_t = \mathcal{U}_{t-1} - \text{pseudo\_labeled} $\;
 Retrain $C$ using $\mathcal{L}_t$ \;
 $V_t: $ Performance of  $C$ on $\mathcal{V}$ \;
}
 \caption{Self-learning}\label{algo:self-learning}
\end{algorithm}

One point illustrated in Algorithm \ref{algo:self-learning} to be highlighted  is that the training set $\mathcal{L}$ is augmented after each iteration of self-learning in a stratified way. In this case, the class distribution of the $N$ pseudo-labeled examples that are added to $\mathcal{L}$ is the same as the class distribution of $\mathcal{L}$. This constraint follows from the independent and identically distributed (i.i.d.) assumption between the $\mathcal{L}$ and $\mathcal{U}$ sets and ensures that the distribution on the classes in the training set does not change as training proceeds. Another point to be mentioned is that the examples that are added to $\mathcal{L}$ may be noisy. Despite the confident predictions of the classifier $C$, one should expect that some of the instances added are wrongly classified. To reduce the impact of the noise to the training set, we monitor the performance of the classifier using the validation set $\mathcal{V}$ and if the performance degrades the self-learning iteration stops.


\section{Experimental Setups}
\label{sect:experimental_setups}

In this section, we present the experimental setups of our study so that our results can easily be reproduced. In particular, we detail the learning frameworks, and the datasets and their splits.

\subsection{Learning Frameworks}
In order to evaluate the effects of our learning strategy, we implement the following baseline systems: (1) Majority Baseline; (2) Logistic Regression that has shown positive results in \cite{Santus2017}, and (3) Feed-forward neural network with two hidden layers of 50 neurons each, which is the direct one-task counterpart of our NN multi-task architecture. For the multi-task learning algorithm, we implemented the architecture shown in Figure \ref{fig:nn-archi} using Keras \cite{chollet2015keras}. In particular, we define 2 fully- connected hidden layers  (i.e. $h_1$, $h_2$, $R=2$) of 50 neurons each. While the number of hidden layers is a free parameter to tune, we select two hidden layers in advance so that the complexity of the multi-task models are comparable to the neural network baseline. The word embeddings are initialized with the 300-dimensional representations of GloVe \cite{pennington2014glove}. The activation function of the hidden layers is the sigmoid function and the weights of the layers are initialized with a uniform distribution scaled as described in \cite{glorot2010understanding}. As for the learning process, we use the Root Mean Square Propagation (RMSprop) optimization method with learning rate set to 0.001 and the default value for $\rho=0.9$. 
For every task, we use the binary cross-entropy loss function. The network is trained with batches of 32 examples\footnote{The code is freely available upon demand for research purposes.}. For the Logistic Regression, we used the implementation of scikit-learn \cite{pedregosa2011scikit}.



\subsection{Datasets and Splits}

\begin{table*}[h]\centering \small
\begin{tabular}{c rrrr rrrr}
\toprule
Dataset & Pairs & $V$ & $V_{\text{train}}/V_{\text{test}}$ & Co-hyp. & Hypernyms & Mero. & Synonyms & Random \\\midrule
ROOT9 & 6,747 &2,373 & 1,423/950& 939/665 & 806/486 & N/A & N/A &339/210\\
RUMEN & 18,979 &9,125 & 5,475/3,650& N/A & 2,638/737 & N/A & 2,256/957&2,227/965\\
ROOT9+RUMEN & 25,726 & 9,779 & 5,867/3,912& 1,193/350 & 3,330/1,238 & N/A &2,297/1,002 &2,630/1,160\\
BLESS & 14,547 & 3,582 & 3,181/2,121 & 1,361/502 & 525/218 & 559/256 & N/A&2,343/971\\
\bottomrule
\end{tabular}
 \caption{\label{datasplit} Statistics on the datasets and the lexical splits we performed to obtain the train and test subsets. $V$ is the vocabulary size in the original dataset; $V_{\text{train}}$ (resp. $V_{\text{test}}$) corresponds to the vocabulary size in the train (resp. test) dataset for the lexical split after removing all words that do not belong to GloVe dictionary. Then, for each lexical relations, we provide the number of word pairs in the train/test datasets. During pre-processing, the train subset has been further split into train, validation and unlabeled sets as explained in section \ref{sect:experimental_setups}.}
\end{table*}

In order to perform our experiments, we use the ROOT9 dataset\footnote{https://github.com/esantus/ROOT9} \cite{Santus2016} that contains 9,600 word pairs, randomly extracted from three well-known datasets: EVALution \cite{santus2015evaluation}, Lenci/Benotto \cite{benotto2015} and BLESS \cite{Baroni2011Bless}. The word pairs are equally distributed among three classes (hypernymy, co-hyponymy and random) and involve several part-of-speech tags (adjectives, nouns and verbs). Here, we exclusively focus on nouns and keep 1,212 hypernyms, 1,604 co-hyponyms and 549 random pairs that can be represented by GloVe  embeddings \cite{pennington2014glove}.

In order to include synonymy as a third studied semantic relation, we build the RUMEN dataset\footnote{The dataset is freely available upon demand for research purposes.} that contains 18,978 word pairs equally organized amongst three classes (hypernymy, synonymy and random). By doing so, we expect to move to more challenging settings and overcome the limitations of only using hypernyms and co-hyponyms  \cite{camacho2017we}. In RUMEN, all noun pairs are randomly selected based on WordNet 3.0\footnote{http://wordnetcode.princeton.edu/3.0/} \cite{Miller1990} such that hypernyms are not necessarily in direct relation and random pairs have as most common parent the root of the hierarchy with a minimum path distance equals to 7\footnote{This value was set experimentally.} to ensure semantic separateness. Finally, we keep 3,375 hypernym, 3,213 synonym and 3,192 random word pairs encoded by GloVe embeddings.

Following a classical learning procedure, the datasets must be split into different subsets: train, validation, test and unlabeled in the case of semi-supervision. The standard procedure is random splitting where word pairs are randomly selected without other constraint to form the subsets. However, \cite{levy2015supervised} point out that using distributional representations in the context of supervised learning tends to perform lexical memorization. In this case, the model mostly learns independent properties of single terms in pairs. For instance, if the training set contains word pairs like ($bike$, $tandem$), ($bike$, {\em off-roader}) and ($bike$, $velocipede$) tagged as hypernyms, the algorithm may learn that $bike$ is a prototypical hypernym and all new pairs ($bike$, $y$) may be classified as hypernyms, regardless of the relation that holds between $bike$ and $y$. To overcome this situation and prevent the model from overfitting by lexical memorization, \cite{levy2015supervised} suggested to split the train and test sets such that each one contains a distinct vocabulary. This procedure is called lexical split. Within the scope of this study, we propose to apply lexical split as defined in \cite{levy2015supervised}. So, lexical repetition exists in the train, validation and the unlabeled subsets, but the test set is  exclusive in terms of vocabulary. Table \ref{datasplit} shows the vocabulary and the pairs before and after the lexical splits. In our experiments, we have further split the pairs dubbed as train so that 60\% of them are unlabeled examples. From the remaining 40\%, we have randomly selected 30\% for validation, resulting in few training examples, which resembles more to a realistic learning scenario where only few positive examples are known. So, while lexical split ensures that the network generalizes to unseen words, it also results in significantly smaller datasets due to the way that these datasets are produced. All subsets are available for replicability\footnote{All datasets are freely available upon demand for research purposes.}.



\section{Results}

In the experiments that follow, we report two evaluation measures: Accuracy and Macro-average F$_1$ measure (MaF$_1$). Accuracy captures the number of correct predictions over the total predictions, while MaF$_1$ evaluates how the model performs across the different relations as it averages the F$_1$ measures of each relation without weighting the number of examples in each case. In the remaining of this section, we comment on three experiments.\\

\begin{table*}[h]\centering
{\small
\begin{tabular}{clcccc|cc}
\toprule
&& \multicolumn{2}{c}{Co-hypo. vs Random} & \multicolumn{2}{c}{Hyper. vs Random} & \multicolumn{2}{c}{Average Results} \\
\cmidrule(rl){3-4}\cmidrule(rl){5-6}\cmidrule(rl){7-8}
&Algorithm & Accuracy & MaF$_{1}$ & Accuracy & MaF$_{1}$ & Accuracy & MaF$_{1}$ \\
\midrule
\parbox[t]{2mm}{\multirow{6}{*}{\rotatebox[origin=c]{90}{ROOT9}}}

&Majority Baseline &  0.760 & 0.431 & 0.698 & 0.411  & 0.729 & 0.421\\
&Logistic Regression & { \bf 0.893} & 0.854 &  0.814 & 0.762 & {\bf 0.854} & 0.808 \\
&NN Baseline & 0.890 & 0.851 & 0.803 & 0.748 & 0.847 & 0.800 \\ \cline{2-8}
&Self-learning &  0.869 &  {\bf 0.859} & 0.816 & 0.772 & 0.843 & {\bf 0.815} \\ 
&Multitask learning & 0.882 & 0.833 & { \bf 0.818} & {\bf 0.773} & 0.850 & 0.803 \\ 
&Multitask learning + Self-learning &
0.854 &   0.811 &  0.810 & 0.767 & 0.832 & 0.789 \\
\midrule
\midrule
&& \multicolumn{2}{c}{Syn. vs Random} & \multicolumn{2}{c}{Hyper. vs Random} & \multicolumn{2}{c}{Average Results} \\
\cmidrule(rl){3-4}\cmidrule(rl){5-6}\cmidrule(rl){7-8}
&Algorithm & Accuracy & MaF$_{1}$ & Accuracy & MaF$_{1}$ & Accuracy & MaF$_{1}$ \\
\midrule
\parbox[t]{2mm}{\multirow{6}{*}{\rotatebox[origin=c]{90}{RUMEN}}}
&Majority Baseline &  0.496 & 0.331 & 0.432 & 0.301 & 0.464 & 0.316\\
&Logistic Regression &  0.628 & 0.628 &  0.711 & 0.706 & 0.670 & 0.667\\
&NN Baseline & 0.679 & 0.678 & 0.752 & 0.748 & 0.716 & 0.713\\ \cline{2-8}
&Self-learning & 0.686 &  0.685 & 0.757 & 0.754 & 0.722 & 0.720\\ 
&Multitask learning & 0.706 & 0.700 & 0.755 &  0.750  & 0.731 & 0.725\\ 
&Multitask learning + Self-learning &
{\bf 0.708} &  {\bf 0.708} & {\bf 0.760} &  {\bf 0.755} & {\bf 0.734} & {\bf 0.732}\\

\bottomrule
\end{tabular}
\caption{\label{resultsRoot9Rumen} Accuracy and Macro F$_1$ scores on ROOT9 and RUMEN datasets for GloVe semantic space.}
}
\end{table*}

In the {\bf first experiment}, we propose to study the impact of the concurrent learning of co-hyponymy (bike $\leftrightarrow$ scooter) and hypernymy (bike $\rightarrow$ tandem) following the first findings of \cite{Yu2015}. For that purpose, we propose to apply our (semi-supervised) multi-task learning strategy over the lexically split ROOT9 dataset using vector concatenation of GloVe \cite{pennington2014glove} as feature representation. Results are illustrated in Table \ref{resultsRoot9Rumen}. The multi-task paradigm shows that an improved MaF$_1$ score can be achieved by concurrent learning without semi-supervision achieving a value of 77.3\% (maximum value overall). In this case, a 1.1\% improvement is obtained over the best baseline (i.e. logistic regression) for hypernymy classification, indeed suggesting that there exists a learning link between hypernymy and co-hyponymy. However, the results for co-hyponymy classification can not compete with a classical supervised strategy using logistic regression. In this case, a 2.1\% decrease in MaF$_1$ is evidenced suggesting that the gains for hypernymy classification are not positively balanced by the performance of co-hyponymy. So, we can expect an improvement for hypernymy classification but not for co-hyponymy in a multi-task environment, suggesting a positive influence of co-hyponymy learning towards hypernymy but not the opposite. Interestingly, the results of the semi-supervised strategy reach comparable figures compared to the multi-task proposal (even superior in some cases), but do not complement each other for the semi-supervised multi-task experiment. In this case, worst results are obtained for both classification tasks suggesting that the multi-task model is not able to correctly generalize from a large number of unlabeled examples, while this is the case for the one-task architecture. 

In the {\bf second experiment}, we propose to study the impact of the concurrent learning of synonymy (bike $\leftrightarrow$ bicycle) and hypernymy following the experiments of \cite{Santus2017} which suggest that symmetric similarity measures (usually tuned to detect synonymy \cite{Kiela2015}) improve hypernymy classification. For that purpose, we propose to apply the same models over the lexically split RUMEN dataset. Results are illustrated in Table \ref{resultsRoot9Rumen}. The best configuration is the combination of multi-task learning with self-learning achieving maximum accuracy and MaF$_1$ scores for both tasks. The improvement equals to 0.7\% in terms of MaF$_1$ for hypernymy and reaches 3\% in terms of MaF$_1$ for synonymy when compared to the best baseline (i.e. neural network). The overall average improvement (i.e. both tasks combined\footnote{Column 3 of Table \ref{resultsRoot9Rumen}.}) reaches 1.8\% for accuracy and 1.9\% for MaF$_1$ over the best baseline. So, these results tend to suggest that synonymy identification may greatly be impacted by the concurrent learning of hypernymy and vice versa (although to a less extent). In fact, these results consistently build upon the positive results of the multi-task strategy without semi-supervision and the self-learning approach alone that both improve over the best baseline results. Note that the results obtained over the RUMEN dataset by the baseline classifiers are lower than the ones reached over ROOT9 for hypernymy, certainly due to the complexity of the datasets themselves. So, we may hypothesize that the multi-task strategy plays an important role by acting as a regularization process and helping in solving learning ambiguities, and reaches improved results over the one-task classifiers. 

In the {\bf third experiment}, we propose to study the impact of the concurrent learning of co-hyponymy, synonymy and hypernymy all together. The idea is to understand the inter-relation between these three semantic relations that form the backbone of any taxonomic structure. For that purpose, we propose to apply the models proposed in this paper over the lexically split ROOT9+RUMEN dataset\footnote{Note that due to the lexical split process, results can not directly be compared to the ones obtained over ROOT9 or RUMEN.}. Results are illustrated in Table \ref{resultsRoot9RumenBless}. The best configuration for all the tasks combined (i.e. co-hyponymy, synonymy and hypernymy) is multi-task learning without semi-supervision. Overall, improvements up to 1.4\% in terms of accuracy and 2\% in terms of MaF$_1$ can be reached over the best baseline (i.e. neural network). In particular, the MaF$_1$ score increases 4.4\% with the multi-task strategy without self-learning for co-hyponymy, while the best result for synonymy is obtained by the semi-supervised multi-task strategy with an improvement of 1.1\% MaF$_1$ score. The best configuration for hypernymy is evidenced by self-learning alone, closely followed by the multi-task model, reaching improvements in MaF$_1$ scores of 1.7\% (resp. 1\%) for self-learning (resp. multi-task learning). Comparatively to the first experiment, both learning paradigms (i.e. semi-supervision and multi-task) tend to produce competitive results alone, both exceeding results of the best baseline. However, the multi-task model hardly generalizes from the set of unlabeled examples, being synonymy the only exception. Finally, note that co-hyponymy seems to be the simplest task to solve, while synonymy is the most difficult one, over all experiments.   


\section{Studying Meronymy}

\begin{table*}[h]\centering
{\small
\begin{tabular}{clcccccc|cc}
\toprule
&& \multicolumn{2}{c}{Co-hypo. vs Random} & \multicolumn{2}{c}{Hyper. vs Random} & \multicolumn{2}{c}{Syn. vs Random} & \multicolumn{2}{c}{Average Results} \\
\cmidrule(rl){3-4}\cmidrule(rl){5-6}\cmidrule(rl){7-8}\cmidrule(rl){9-10}
&System & Accuracy & MaF$_{1}$ & Accuracy & MaF$_{1}$  & Accuracy & MaF$_{1}$ & Accuracy & MaF$_{1}$\\
\midrule
\parbox[t]{2mm}{\multirow{6}{*}{\rotatebox[origin=c]{90}{ROOT+RUMEN}}}
&Majority Baseline &  0.768 & 0.434 & 0.516 & 0.340  & 0.536 & 0.350  & 0.607& 0.375\\
&Logistic Regression &  0.909 & 0.872 &  0.669 & 0.669& 0.634 & 0.632  & 0.737& 0.724\\
&NN Baseline & 0.914 & 0.875 & 0.712 & 0.712 & 0.663 & 0.659  & 0.763& 0.748\\ \cline{2-10}
&Self-learning & 0.928 &  0.900 &  \textbf{0.729} &  \textbf{0.729} & 0.668 & 0.665  & 0.775& 0.765\\ 
&Multitask learning & \textbf{0.943} &  \textbf{0.919} & 0.723 &  0.722  & 0.666 & 0.664  & {\bf 0.777}& {\bf 0.768}\\ 
&Multitask learning + Self. &
0.939 &   0.911 & 0.711 & 0.711 &  \textbf{0.672} &  \textbf{0.670}  & 0.774& 0.764\\
\midrule
\midrule
&& \multicolumn{2}{c}{Co-hypo. vs Random} & \multicolumn{2}{c}{Hyper. vs Random} & \multicolumn{2}{c}{Mero. vs Random} & \multicolumn{2}{c}{Average Results}\\
\cmidrule(rl){3-4}\cmidrule(rl){5-6}\cmidrule(rl){7-8}\cmidrule(rl){9-10}
&System & Accuracy & MaF$_{1}$ & Accuracy & MaF$_{1}$  & Accuracy & MaF$_{1}$ & Accuracy & MaF$_{1}$\\
\midrule
\parbox[t]{2mm}{\multirow{6}{*}{\rotatebox[origin=c]{90}{BLESS}}}
&Majority Baseline &  0.660 & 0.397 & 0.816 & 0.449 & 0.653 & 0.395  & 0.710& 0.414\\
&Logistic Regression & 0.845 & 0.830 & 0.888 & 0.794 & 0.748 & 0.723   & 0.827& 0.782\\
&NN Baseline & 0.870 & 0.855 & 0.892 & 0.809 & 0.738 & 0.709  & 0.833& 0.791\\ \cline{2-10}
&Self-learning &   0.877 & {\bf 0.863} & 0.900 & 0.807 & 0.749 & 0.723  & 0.842& 0.798\\ 
&Multitask learning & 0.866 & 0.847 & {\bf 0.903} & {\bf 0.816} & {\bf 0.764} & {\bf 0.733}  & {\bf 0.844} & 0.799 \\ 
&Multitask learning + Self. & {\bf 0.878} & {\bf 0.863} & 0.900 & 0.813 & 0.754 & {\bf 0.733}  & {\bf 0.844} & {\bf 0.803}\\

\bottomrule
\end{tabular}
\caption{\label{resultsRoot9RumenBless}Accuracy and Macro F$_1$ scores on ROOT9+RUMEN and BLESS datasets for GloVe semantic space.}
}
\end{table*}

In this section, we study the introduction of the meronymy relation (bike $\rightarrow$ chain) into a multi-task environment, as it has traditionally been studied together with hypernymy \cite{Glavas2017}. The overall idea is to verify whether the meronymy semantic relation can benefit from the concurrent learning of the backbone semantic relations that form knowledge bases. For that purpose, we apply our learning models over the lexically split BLESS dataset \cite{Baroni2011Bless} that includes three semantic relations: co-hyponymy, hypernymy and meronymy. The details of the lexical split is presented in Table \ref{datasplit} and note that the BLESS dataset has been processed in the exact same way as ROOT9 and RUMEN, i.e. retaining only noun categories and word pairs that can be represented by the GloVe semantic space. Results are presented in Table \ref{resultsRoot9RumenBless}. The best configuration over the three tasks combined is obtained by the semi-supervised multi-task strategy with a MaF$_1$ score equals to 80.3\%, thus improving 1.2\% over the best baseline (i.e. neural network). In particular, we can notice that the most important improvement is obtained for the meronymy relation that reaches 73.3\% for MaF$_1$ and 76.4\% for accuracy with the multi-task model without semi-supervision. In this particular case, the improvement is up to 2.6\% in accuracy and 2.4\% in MaF$_1$ over the neural network baseline. For co-hyponymy (resp. hypernymy), best results are obtained by multi-task with semi-supervision (resp. without semi-supervision), but show limited improvements over the best baseline, suggesting that meronymy gains more in performance from the concurrent learning of co-hyponymy and hypernymy than the contrary, although improvements are obtained in all cases. Comparatively to the other experiments, we also notice that although the self-learning algorithm and the multi-task framework without semi-supervision perform well alone, the combination of both strategies does not necessary lead to the best results overall, suggesting that the present architecture can be improved to positively gain from the massive extraction of unlabeled examples.    


\section{Conclusions and Discussion}

In this paper, we proposed to study the concurrent learning of semantic relations (co-hyponymy, hypernymy, synonymy and meronymy) using simple learning strategies (self-learning and hard parameter sharing multi-task learning) and state-of-the-art continuous distributional representations (concatenation of GloVe embeddings). The idea was to verify if the concurrent learning of these relations that share some cognitive similarities could be beneficial, without necessarily focusing on overall performance. Obtained results show that concurrent learning can lead to improvements in a vast majority of tested situations. In particular, we have shown that within this experimental framework, hypernymy can gain from co-hyponymy, synonymy from hypernymy, co-hyponymy from both hypernymy and synonymy, and meronymy from both co-hyponymy and hypernymy. Moreover, it is interesting to notice that in three cases out of four, the improvement obtained by the multi-task strategy is obtained for the most difficult task to handle, thus suggesting the benefits of concurrent learning. Based on these preliminary findings, a vast amount of improvements can now be introduced into the framework to increase overall performance. With respect to the input features, we intend to study the potential benefits from dedicated embeddings such as hypervec \cite{Nguyen2017}, knowledge graphs embeddings \cite{speer2017conceptnet} and dual embeddings \cite{Nalisnick2016}. Moreover, we deeply believe that the LSTM path-based features introduced in \cite{Shwartz2016} and some well-defined word pairs similarity measures \cite{Santus2017} can lead to classification improvements by complementing the information present in distributional semantic spaces. With respect to the learning framework, some clear efforts must be performed. Indeed, the current combination of self-learning and hard parameter sharing multi-task learning is not beneficial in a vast majority of cases suggesting the proposition of new architectures following the ideas of Tri-training \cite{ruder2018}. Moreover, it is clear that more complex architectures such as convolutional neural networks may improve the learning process as it is proposed in \cite{Attia2016} for similar tasks. As semi-supervision is concerned, we intend to study a more realistic situation where unlabeled examples are massively gathered by lexico-syntactic patterns \cite{Hearst1992} or by paraphrase alignment \cite{dias2010} over huge raw text corpora. Indeed, here, semi-supervision is just simulated by fractioning the existing datasets. Finally, we plan to difficult the original task by including the detection of the direction of the asymmetric semantic relations, testing the combination of more closely related semantic relations and including noisy pairs as in \cite{Vylomova2016}. 

\balance
\bibliographystyle{acl_natbib_nourl}
\bibliography{emnlp2018}

\end{document}